%% file: main.tex
\documentclass[letterpaper, 10 pt, conference]{ieeeconf}    
\IEEEoverridecommandlockouts
\input{./meta/packages}
\begin{document}
\def \TITLE {Vision-based Oxy-fuel Torch Control for Robotic Metal Cutting}
\def \TITLEheader {\def\\{\relax\ifhmode\unskip\fi\space\ignorespaces}\TITLE}

\title{\LARGE \bf \TITLE}
\author{James~Akl, Yash~Patil, Chinmay~Todankar, and~Berk~Calli%
        \thanks{The authors are with the Robotics Engineering Department, Worcester~Polytechnic~Institute, 27 Boynton St, Worcester, MA 01609, USA.}%
        \thanks{E-mail: $\mathtt{\{jgakl, yrpatil, ctodankar, bcalli\}@wpi.edu}$}%
}
\markboth{IEEE ROBOTICS AND AUTOMATION LETTERS}{Akl \MakeLowercase{\textit{et al.}}: \TITLEheader}
\maketitle

\begin{abstract} 
    The automation of key processes in metal cutting would substantially benefit many industries such as manufacturing and metal recycling.
    We present a vision-based control scheme for automated metal cutting with oxy-fuel torches, an established cutting medium in industry.
    The system consists of a robot equipped with a cutting torch and an eye-in-hand camera observing the scene behind a tinted visor.
    We develop a vision-based control algorithm to servo the torch's motion by visually observing its effects on the metal surface.
    As such, the vision system processes the metal surface's heat pool and computes its associated features, specifically pool convexity and intensity, which are then used for control.
    The operating conditions of the control problem are defined within which the stability is proven.
    In addition, metal cutting experiments are performed using a physical 1-DOF robot and oxy-fuel cutting equipment.
    Our results demonstrate the successful cutting of metal plates across three different plate thicknesses, relying purely on visual information without \textit{a priori} knowledge of the thicknesses.
\end{abstract}


\IEEEpeerreviewmaketitle

\section{Introduction} \label{sec:introduction}
    \input{./sections/1_Introduction}
\section{Related Work} \label{sec:related}
    \input{./sections/2_Related}
\section{Cutting Problem Formulation} \label{sec:problem}
    \input{./sections/4_Problem}
\section{Vision Feedback Design} \label{sec:vision}
    \input{./sections/5_Vision}
\section{Control Algorithm Design} \label{sec:control}
    \input{./sections/6_Control}
\section{Physical Cutting Experiments} \label{sec:evaluation}
    \input{./sections/7_Evaluation}
\section{Conclusion} \label{sec:conclusion}
    \input{./sections/8_Conclusion}




\bibliographystyle{IEEEtran}
\bibliography{./meta/references}



\end{document}

%% file: meta/packages.tex
\usepackage{cite}
\usepackage[pdftex]{graphicx}
\usepackage{multirow}
\usepackage{amsmath,amssymb,amsfonts}
\usepackage{listings}
\usepackage[vlined,ruled]{algorithm2e}
\usepackage{caption, subcaption}
\usepackage[normalem]{ulem}

%% file: sections/1_Introduction.tex

\begin{figure}[!t]
  \begin{center}
  \includegraphics[width=0.48\textwidth]{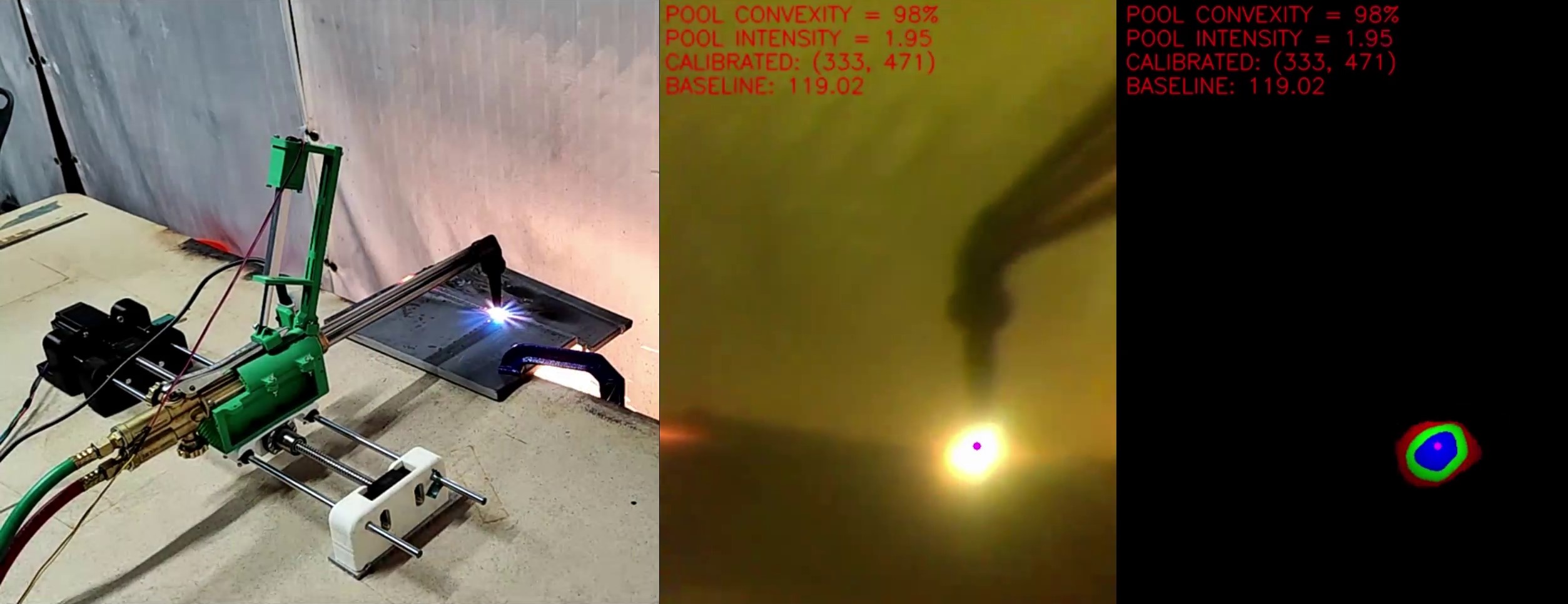}
  \end{center}
  \caption{The proposed vision-based control algorithm enables the 1-DOF cutting robot to autonomously cut steel plates of different thicknesses.
  \textbf{Left:} Third-person view of the robot during control.
  \textbf{Center:} Eye-in-hand view of the metal surface showing the heat pool.
  \textbf{Right:} Processed stream of the eye-in-hand footage focusing on the shown heat pool.
  }
  \label{fig:montage}
 \vspace{-12pt}
\end{figure}

The use of robots in industrial applications contributes significantly to productivity growth in the aggregate economy~\cite{Graetz2018}.
Furthermore, increased adoption of industrial robots is shown to improve worker safety \cite{Gihleb2022} and reduce the global ecological footprint~\cite{Chen2022}.
As the demand for major metals is projected to increase~\cite{Watari2021}, their associated industrial operations become ever more important.
In particular, the automation of metal cutting can substantially benefit cutting-intensive sectors such as manufacturing and metal scrap recycling—the latter of which includes ship-breaking (breaking large metal structures down to smaller fragments) and results in more energy-efficient steel production as compared to processing iron ores~\cite{Harvey2021}.
Metal cutting technologies and their associated robots are diverse and carry distinct advantages and limitations~\cite{Bogue2008}.
The focus of this work is to automate oxy-fuel cutting operations, which use a combustion-based thermal cutting medium (see Fig.~\ref{fig:montage}).

\begin{figure*}[!htb]
  \begin{center}
  \includegraphics[width=0.98\textwidth]{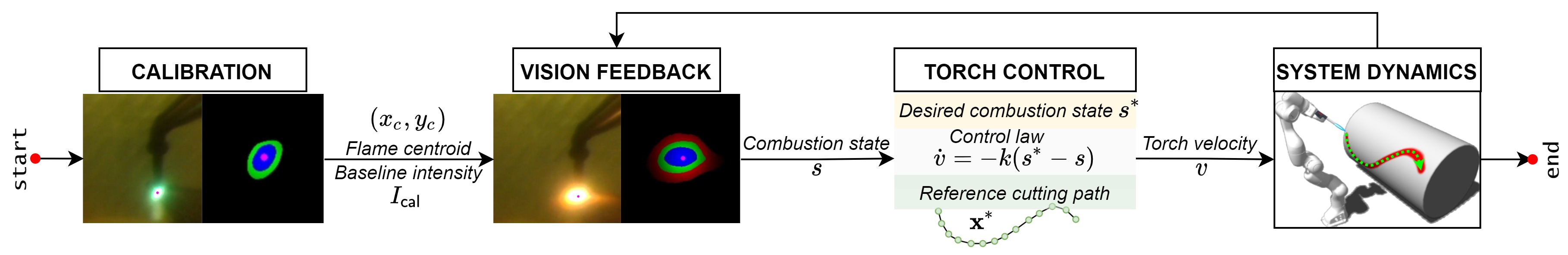}
  \end{center}
  \caption{Overview of the proposed method.
  The vision system is calibrated autonomously to provide vision feedback used for controlling the torch velocity along the reference cutting path.
  Generating the path is addressed in our prior work \cite{Akl2023}.}
  \label{fig:overview}
 \vspace{-12pt}
\end{figure*}

We develop a vision-based control algorithm (illustrated in Fig.~\ref{fig:overview}) to perform autonomous combustion cutting using an oxy-fuel torch.
This approach is inspired by the techniques of skilled cutting workers, particularly in ship-breaking yards that we surveyed.
The workers track the formation and evolution of a \textit{heat pool} on the metal surface to adjust their torch motions for successful cutting.
In the same spirit, we design a vision system that encodes the heat pool's visual characteristics (shape, size, brightness, and color) by computing two features: the heat pool's convexity and intensity.
These two features are combined to describe the heat pool's combustion state, which is utilized for controlling the torch's motion.
We formalize the oxy-fuel cutting problem into a sequence of tasks and specify the corresponding robotic cutting system's core components.
We implement our vision-based control algorithm for a 1-DOF cutting robot whose setup, camera view, and vision feedback are shown in Fig.~\ref{fig:montage}.

The core contributions of this work are:
\begin{itemize}
    \item Formulating the oxy-fuel cutting problem from the robotic automation standpoint;
    \item Developing a novel vision-based control algorithm for combustion cutting; and
    \item Experimentally evaluating our algorithm on a 1-DOF robot cutting steel plates.
\end{itemize}

In this work, our formulation decouples the torch control problem from the manipulator's motion planning and control tasks.
We achieve this by formulating and solving the torch control problem in the task space.
In effect, we assume that:
\begin{enumerate}
    \item The cutting path is predetermined by an appropriate cutting path generation approach, \textit{e.g.}, our prior work~\cite{Akl2023}.
    \item The robot uses conventional motion planners and controllers to keep the torch orthogonal to the metal surface while traversing the cutting path.
\end{enumerate}
As such, the 3D cutting path of the robot is assumed given, and the purpose of our vision-based control algorithm is then to determine the cutting speed, \textit{i.e.}, the torch flame's tangential velocity at which the robot traverses the cutting path (for illustration, refer to Fig.~\ref{fig:cutting_formulation}).
Crucially, this cutting speed determines the success of the cutting operation.
As also demonstrated in our experiments, setting a fixed cutting speed does not guarantee successful cuts, since the speed must be regulated per the surface's combustion conditions (as is practiced by skilled metal cutting workers).

\uline{To the best of our knowledge, this work presents the first method in the literature to close this loop via a vision-based controller that regulates the combustion state by updating the torch speed}.
We provide a stability analysis of our controller and experimentally show that the method can successfully cut steel plates of different thicknesses relying only on visual feedback without \textit{a priori} knowledge about the thicknesses.

%% file: sections/2_Related.tex
This section reviews existing work in automated metal cutting and welding, and the use of visual sensing therein.

\subsection{Automated Metal Cutting}
The automation of metal cutting operations is explored in the context of laser cutting~\cite{Xia2016, Ghany2006}, arc cutting~\cite{Bach1996}, machining~\cite{Denkena2015, Iglesias2015}, custom-tool mechanical cutting~\cite{Shin2018}, plasma cutting~\cite{Park2021, Mustafa2016, Wang2001}, and hybrid induction plasma cutting~\cite{Arner2016}.
In comparison, automated oxy-fuel cutting, which is the focus of this work, is sparsely covered in the literature.
The most pertinent of these is~\cite{Yoo2008}, which designs a reactive controller for a gas cutting robot that trims the undesirable strips from conveyor-fed sheet metal.
This control architecture features extensive instrumentation adapted to its specialized strip-cutting application, yet does not use vision-based sensing in more general cutting scenarios.
More traditional CNC-based approaches~\cite{Dandgawhal2020, Kulkarni2008} require the prior specification of task parameters, including cutting speeds.
Thus far, the automation of oxy-fuel cutting is limited to narrow settings or to full specification of task parameters.
However, there are considerable advantages to further the automation of oxy-fuel cutting as covered in~\cite{Nachbargauer2019}.
We believe that our contribution is a significant step towards this goal.

To the best of our knowledge, this is the first work on general-purpose automated oxy-fuel cutting whose formulation is independent of robot geometry and does not require \textit{a~priori} specification of the metal thickness or cutting speeds.

\subsection{Automated Metal Welding}
A related operation to metal cutting is welding whose automation extensively uses vision-based sensing due to being non-invasive, precise, simple, and inexpensive.
\cite{Xu2021}~reviews recent advancements in vision-based automated welding such as visual calibration, seam tracking, weld pool monitoring and deformation detection.

There is extensive work on the visual processing of weld seams and weld pools for improving welding performance.
\cite{Muhammad2017}~develops an active vision system wherein image processing techniques extract welding seam features for vision-based welding control.
In~\cite{He2016}, a feature point extraction scheme for the weld seam profile enables automatic multi-pass route planning for the metal active gas (MAG) arc robotic welding of thick plates.
Seam tracking for MAG welding is also covered in~\cite{Ye2013} by denoising the image via optical filters tailored to the welding process’s light spectrum for improving the seam tracking precision and its stability.
In~\cite{Xu2012}, a vision system acquires clear and steady images during welding using a specialized Canny edge detection algorithm for real-time extraction of welding parameters from the weld seam and weld pool.
\cite{Gao2011}~uses the weld pool’s centroid to improve the weld seam tracking’s real-time accuracy.
\cite{Shen2008}~develops a calibration-free vision-based method for seam gap measuring and seam tracking, and a controller to regulate the weld formation, welding current, and wire feed rate.

Weld pool features extracted from imaging are often used to improve welding performance.
For example, \cite{Wu2021}~uses X-ray imaging to analyze the weld pool flow patterns by tracking particles in order to enhance its flow and the predictability of the welded component’s microstructures, which affect its end-use quality.
\cite{Liu2017}~uses active contours for weld pool boundary composition to overcome the interference of arc light and spatters during welding.
In tungsten inert-gas (TIG) welding, \cite{Wang2005}~exploits the metal alloy’s characteristics for improved image processing and pattern recognition.
The weld pool’s image is simplified via a weighted median filter, a statistical threshold, and a projection, after which a neural network detects the pool's edges.
For real-time torch position control, \cite{Gao2005}~uses Kalman filtering on weld pool images to adjust the torch’s position and improve seam tracking accuracy.
\cite{Bae2002}~uses a charge-coupled device (CCD) camera’s feedback filtered through specialized lenses and processed for use in control.

While there are extensive vision-based methodologies for robotic welding and its control, there is sparse development in robotic oxy-fuel cutting, which this work addresses.

%% file: sections/4_Problem.tex
The control problem consists of moving the torch’s flame jet along a reference spatial path on a metal surface, such that the cut is successful and efficient.
The controller must regulate the surface heat pool's combustion state while traversing the cutting path by moving the torch tip at an appropriate velocity.

We assume that this cutting path is predetermined by a suitable cutting path generation method and that the torch tip is kept normal to the metal surface while moved along this path.
For instance, these planning tasks are achieved in our prior work~\cite{Akl2023} via active vision techniques to obtain the desired cutting path and the surrounding surface normal estimates on the metal surface.
In this work, we focus on determining the cutting speed, \textit{i.e.}, the velocity at which the torch is moved along this path.
Moving the torch at adequate cutting speeds is essential for successful cutting.
In effect, our experiments in section~\ref{sec:evaluation} provide examples (see Fig.~\ref{fig:cuts}) illustrating the outcomes of inappropriate cutting speeds on the metal surface.
We emphasize that our vision-based control algorithm does not use any \textit{a priori} information about the plate thickness or temperature and relies purely on visual feedback.

\begin{figure}[t]
  \begin{center}
  \includegraphics[width=0.48\textwidth]{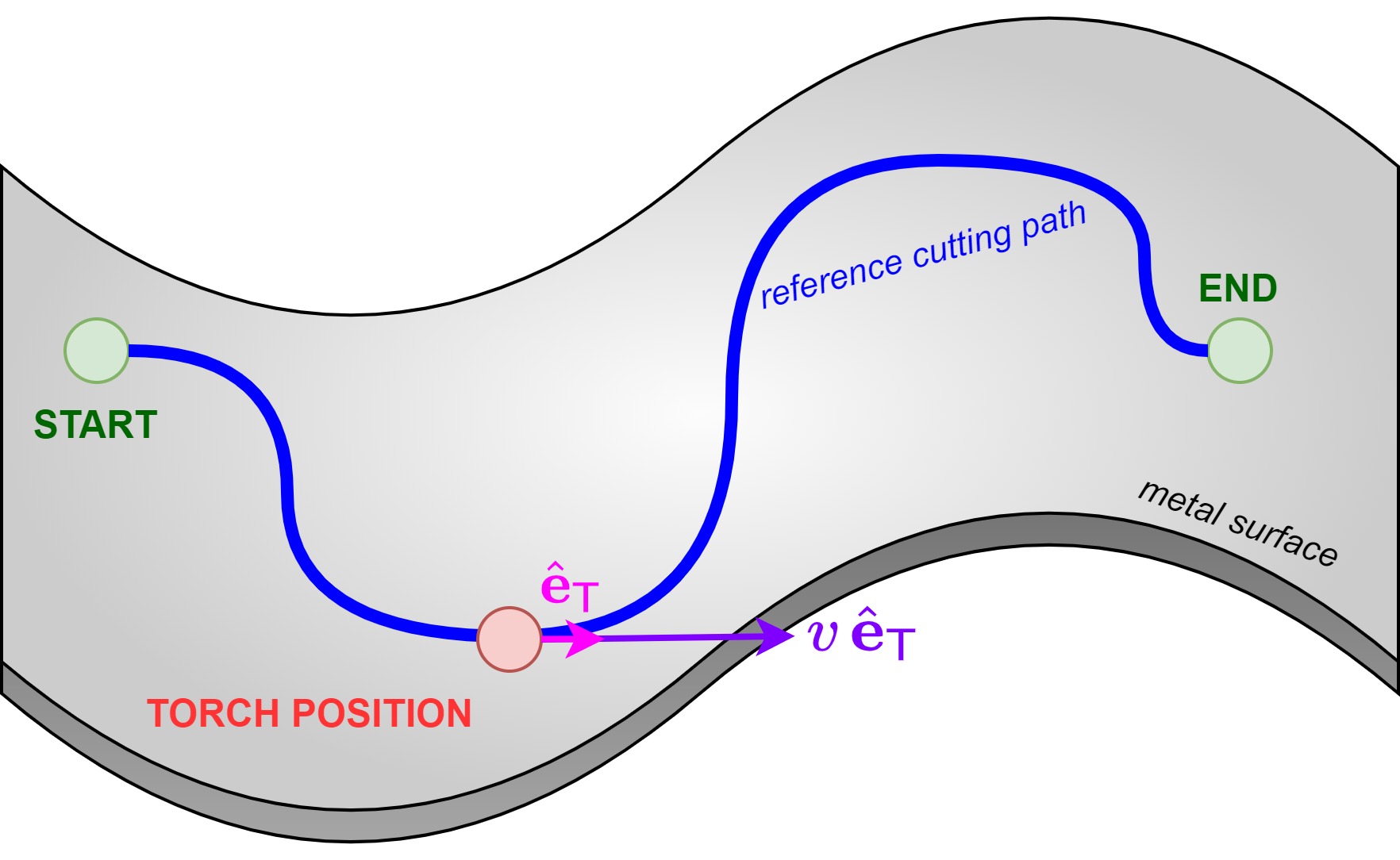}
  \end{center}
  \caption{Illustration of the cutting control formulation showing the tangential motion along the reference path.}
  \label{fig:cutting_formulation}
 \vspace{-12pt}
\end{figure}

Since the desired torch tip positions are known (the points on the reference cutting path) then the controller is constrained to moving the torch tip only along the tangent vector of the curve.
Essentially, the reference path is a sequence of torch tip positions to visit wherein the controller determines the rate at which to traverse these positions, expressed as the velocity along the tangent vector (see Fig.~\ref{fig:cutting_formulation}).

The controller sets the acceleration along the path's tangent vector (updating the velocity) using visual feedback from the RGB camera.
Expressing the control action along the local tangent vector decouples the combustion control problem from the robot motion planning problem.
In general settings, the tangential motion commands can be resolved into joint space via the system’s inverse kinematics.
As such, in our experiments, we use a 1-DOF robotic system, for which the tangent vector coincides with the robot motion, enabling us to focus on combustion control in isolation.

\begin{figure}[t]
  \begin{center}
  \includegraphics[width=0.48\textwidth]{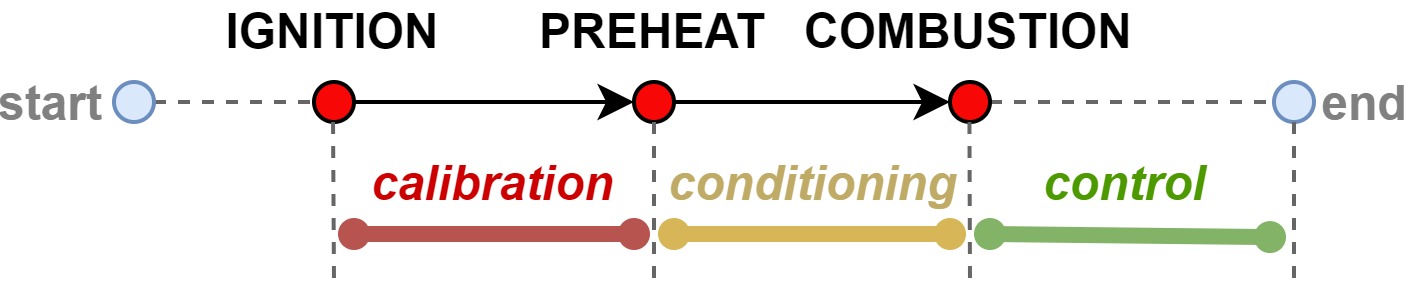}
  \end{center}
  \caption{The oxy-fuel cutting problem formulated as a sequence of events (ignition, preheating, combustion) and tasks (calibration, conditioning, control).}
  \label{fig:cutting_timeline}
 \vspace{-12pt}
\end{figure}

The oxy-fuel cutting problem can be formulated as a sequence of the following events (see Fig.~\ref{fig:cutting_timeline}):
(1)~\textit{Ignition}:~The torch flame is ignited, and the oxy-fuel pressures are adjusted.
(2)~\textit{Preheating}:~The torch flame heats up the metal surface at the initial cutting position.
(3)~\textit{Combustion}:~When the surface is sufficiently hot, the torch's oxygen bypass lever is engaged to increase oxygen flow and intensify the metal's combustion resulting in material removal (cutting).

Accordingly, the cutting system performs these tasks:
\begin{enumerate}
    \item \textit{Calibration:}~Detect the torch flame and record its centroid (by averaging centroid coordinates of the blue region for a stable torch flame across a set number of frames) and its intensity (defined in section~\ref{subsec:heat_pool_intensity}) to be used as a baseline for subsequent measurements.
    \item \textit{Conditioning:}~Maintain the torch flame at the initial position on the cutting path to heat the metal surface.
    As the heat pool forms and evolves, monitor the pool until it is adequate for combustion.
    \item \textit{Control:}~Start the combustion by engaging the oxygen bypass lever and traverse the cutting path by regulating the velocity to maintain the desired pool combustion.
\end{enumerate}

In this work, we address the control task and the vision system's calibration.
During experiments, the torch's ignition is performed manually, and the initiations of calibration, conditioning, and combustion are decided by manual inspection.

%% file: sections/5_Vision.tex
As the purpose of the controller is to maintain a sufficient combustion for cutting, the visual feedback must be designed to describe the heat pool's combustion state.
We note that excessive combustion indicates an overly slow torch velocity, while deficient combustion indicates an overly fast velocity, and an adequate velocity leads to a desired combustion state.
The extracted visual features must then reflect this negative relationship between cutting speed and combustion state.

As such, the visual feedback provided to the controller is a heat pool combustion state descriptor $s$ computed from the camera’s RGB feed.
Specifically, the camera images are processed (simplified and filtered) after which two heat pool features (convexity and intensity) are computed and combined to describe the pool's combustion state $s$.

\subsection{Image Processing}
The image is simplified such that the heat pool's features are emphasized.
The features of interest are visual surrogates for shape and temperature.
In the heat pool, the color distribution of light emissions can be used to estimate a relative temperature profile.
This is because visible light of higher frequencies is emitted at higher energy states.
While color-based temperature estimations are found with varying sophistication~\cite{Lu2009, Yamazaki2010, Wang2019}, we adopt the heuristics that follow.

\begin{figure}[t]
  \begin{center}
  \includegraphics[width=0.48\textwidth]{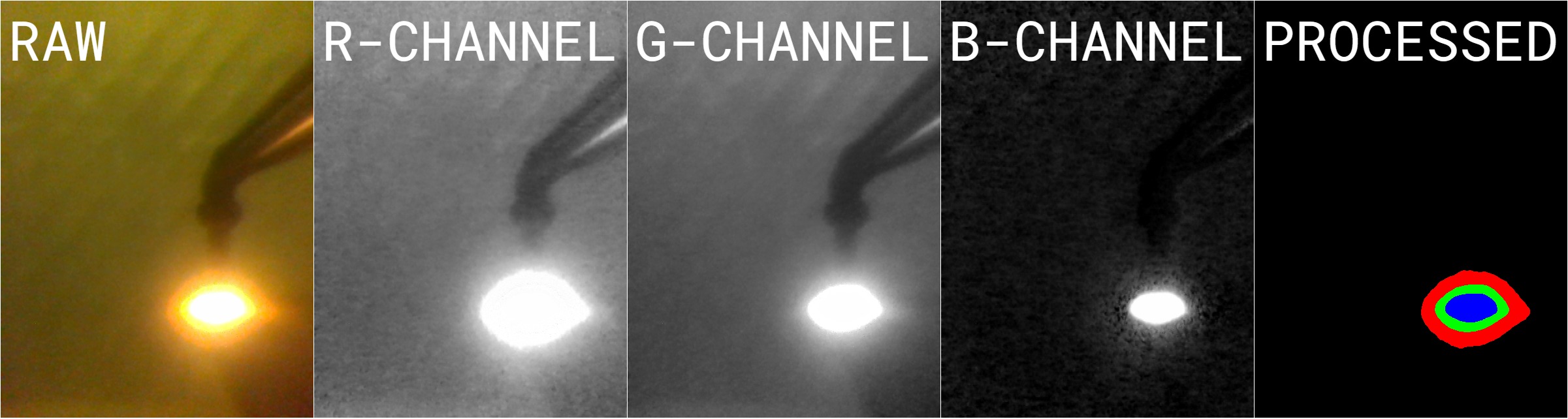}
  \end{center}
  \caption{The raw RGB image from the camera, its red channel, green channel, blue channel, and processed version.}
  \label{fig:color_threshold}
 \vspace{-12pt}
\end{figure}

We restrict the RGB color space to four discrete colors listed in order of decreasing temperature: blue, green, red, and black.
In this discretized color space, blue light emissions correlate with the highest temperatures (higher electromagnetic frequency), green light with moderate temperatures (intermediate frequency), and red light with the lowest temperatures (lower frequency).
Black denotes low light emissions and thus negligible temperatures.
This phenomenon is confirmed in the heat pool footage.
Consider in Fig.~\ref{fig:color_threshold} the raw camera image and its color channels.
The highest heat intensity is nearest to the flame center, and decays with distance.
In effect, the blue channel's highest brightness is nearest to the pool center, whereas the green and red channels include moderate and lower temperature portions of the pool.

With this, the raw image is processed as follows:
\begin{enumerate}
    \item Binary-threshold each channel using high cutoffs.
    \item Color each channel's non-zero pixels (red, green, blue).
    \item Eliminate channel overlap via $\mathtt{XOR}$ operations.
    \item Merge the channels, yielding a segmented heat pool.
\end{enumerate}

This simple and efficient channel-wise thresholding procedure emphasizes the pool's geometric and thermal information.
The shape and size of the pool is preserved and its color information is quantized as the image contains only four colors.
This enables efficient representation and tractable reasoning about the heat pool, its features, and its combustion state.

\subsection{Heat Pool Convexity}
\begin{figure}[t]
  \begin{center}
  \includegraphics[width=0.48\textwidth]{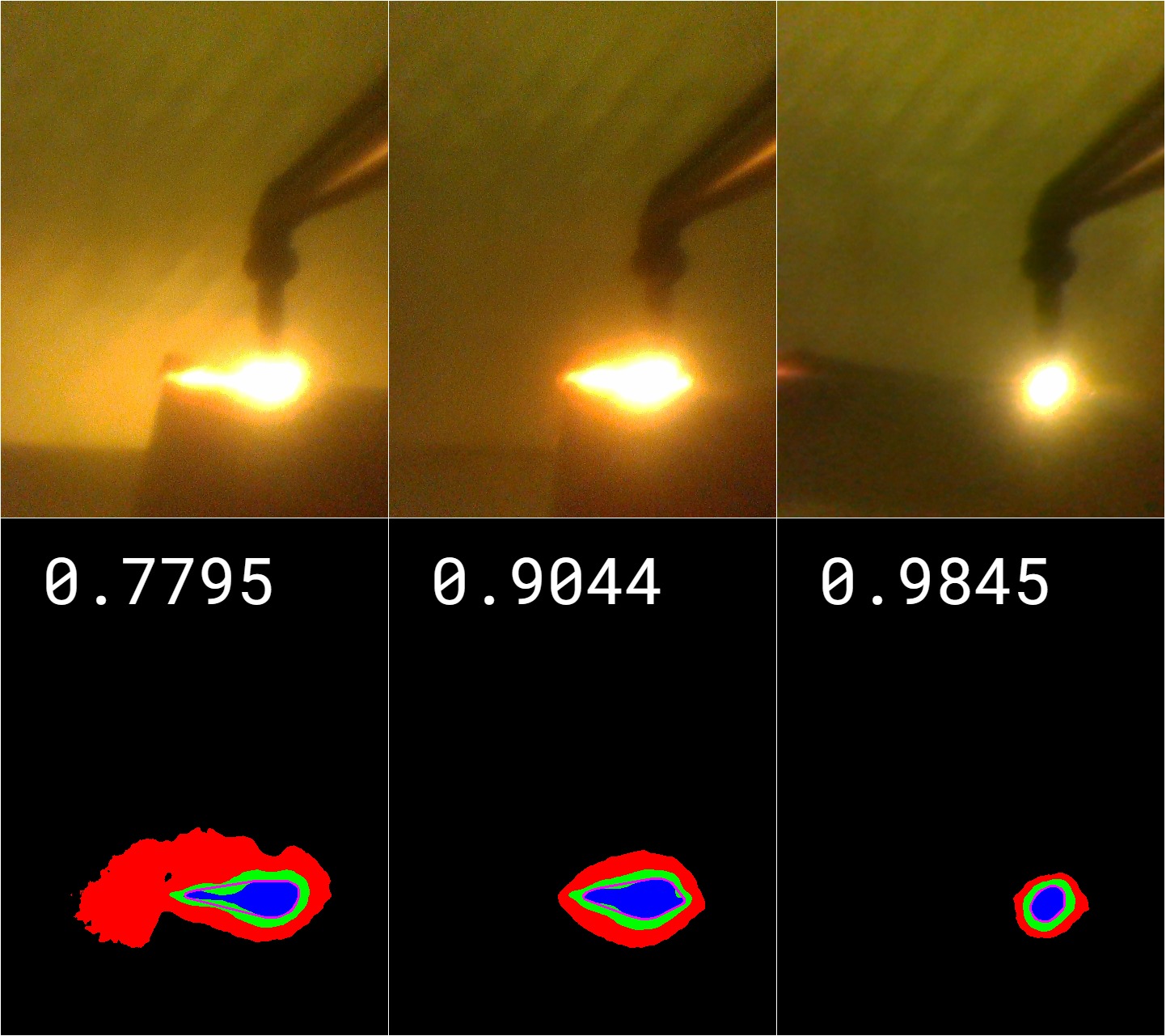}
  \end{center}
  \caption{Demonstrating the pool convexity value (in white) along with the blue layer's convex hull (in magenta).
  \textbf{Left:} The pool has significant convexity defects (deviations from its convex hull), reflected in its low pool convexity of $\mathtt{0.7795}$.
  \textbf{Center:} The pool has moderate convexity defects and has a pool convexity of $\mathtt{0.9044}$.
  \textbf{Right:} The pool has negligible convexity defects and has a high pool convexity of $\mathtt{0.9845}$.
  }
  \label{fig:convexity_defects}
 \vspace{-12pt}
\end{figure}

Utilizing the processed image, we design a measure of the pool's shape using its convexity.
During cutting, when the pool's shape is relatively more convex, then the surface conditions are more adequate for combustion.
Conversely, when the pool's shape is more concave or exhibits significant convexity defects (see Fig.~\ref{fig:convexity_defects}), then the surface conditions are more adverse for combustion.
This is since a higher relative convexity is indicative of heat that is more concentrated in the heat pool.
While there are sophisticated approaches to quantify a closed contour's convexity~\cite{Corcoran2011}, we compute the area ratio between the pool contour against its convex hull.

Formally, let $K$ be the contour of the heat pool in the image; this is a set of pixel coordinates of the simple closed curve.
We define the pool convexity as $c = |K|/|\operatorname{conv} K|$ where $|K|$ denotes the area of the contour and $|\operatorname{conv} K|$ the area of its convex hull.
This yields the following desirable properties:
(1) $c \in (0,1]$ where $c = 1$ is a perfectly convex shape.
(2) A higher value of $c$ denotes a higher convexity and thus correlates with a higher combustion state.
(3) The ratio $c$ is invariant to pool's size or color, it only quantifies its shape.
More precisely, it is dimension-invariant, pose-invariant, and color-invariant.

In practice, restricting the contour $K$ to the hottest layer in the pool, \textit{i.e.}, the pool’s blue channel, yields more robust convexity values.
Moreover, $K$ is retrieved computationally as the largest contour closest to the torch flame centroid (obtained previously from the calibration step).

\subsection{Heat Pool Intensity}\label{subsec:heat_pool_intensity}
The pool intensity is designed to convey, from the processed image, the pool's color and size.
The pool intensity computation is summarized below and elaborated thereafter:
\begin{enumerate}
    \item Weigh pixels based on color (black, red, green, blue).
    \item Weigh pixels via a radial decay function centered at the torch flame's centroid (obtained from calibration).
    \item Sum all weighted pixels to yield an unscaled intensity.
    \item Scale the sum by the calibration baseline, apply a saturation cutoff, and normalize to yield the pool intensity.
\end{enumerate}

\begin{figure}[!t]
  \begin{center}
  \includegraphics[width=0.48\textwidth]{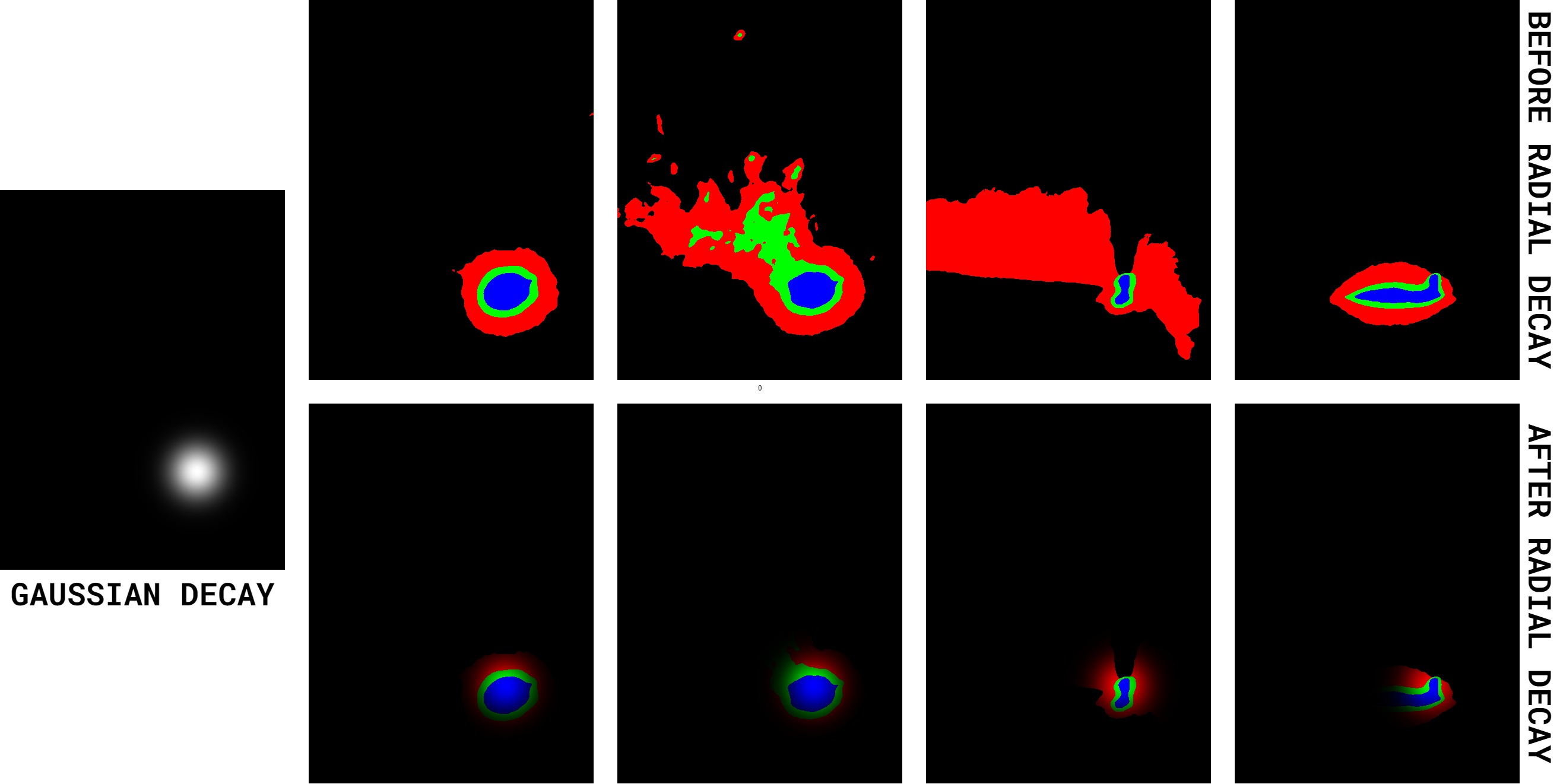}
  \end{center}
  \caption{The Gaussian radial decay (leftmost image) is centered at the torch flame’s centroid.
  In the various scenarios shown (top images), the decay reduces unwanted effects and noise (bottom images), as well as scales the intensities based on the distance from the torch flame's centroid.
  }
  \label{fig:gaussian_weights}
 \vspace{-12pt}
\end{figure}

The first step approximates the relative temperature differences in the quantized color space.
This captures size (total non-zero pixels) and color (via weights).
The second step approximates the nonlinear radial decay effects of heat transfer, \textit{i.e.}, the heat at a pixel decays quickly with distance from the torch flame’s centroid.
Moreover, this radial decay provides robustness against noise and undesirable effects such as: (1) light pollution, (2) sudden sparks, slag, or streaks, and (3) residual heated regions away from the heat pool.
Here, a bivariate Gaussian function is appropriate as it models radial exponential decay using simple parameters.
We observe the effect of weighing via Gaussian decay in Fig.~\ref{fig:gaussian_weights}.
These modeling choices are heuristic-based yet provide a sufficiently accurate and efficient visual feedback signal for control.

Formally, let $\mathrm{M}$ be the processed image in an array form, which contains pixels $p$ considered tuples $(x_p, y_p, \mathrm{color}(p))$ containing the pixel coordinates and color.
The 2D Gaussian function $g(p)$ assigns a decay factor to each pixel based on its distance from the torch flame centroid $(x_c, y_c)$, which is determined at calibration.
The color-weighing function $w(p)$ assigns a constant weight based on the pixel's color.
The functions $g(p)$ and $w(p)$ are defined as:
\begin{equation}
    g(p) = \operatorname{exp} \left[ -\frac{(x_p - x_c)^2}{2\sigma_X^2} - \frac{(y_p - y_c)^2}{2\sigma_Y^2} \right]
\end{equation}

\begin{equation}
    w(p) = \begin{cases}
    0, & \text{if }\mathrm{color}(p) = \text{Black} \\
    w_\mathsf{R}, & \text{if }\mathrm{color}(p) = \text{Red} \\
    w_\mathsf{G}, & \text{if }\mathrm{color}(p) = \text{Green} \\
    w_\mathsf{B}, & \text{if }\mathrm{color}(p) = \text{Blue}
    \end{cases}
\end{equation}

Here, $\sigma_X$ and $\sigma_Y$ determine the radial decay rate in each axis.
The weights $w_\mathsf{R} < w_\mathsf{G} < w_\mathsf{B}$ are constants.
Using these definitions, an absolute sum is defined as $I~=~\sum_{p \in \mathrm{M}} g(p)w(p)$.
In our case $\sigma_X = \sigma_Y = 30\,\mathrm{px}$ and $(w_\mathsf{R},w_\mathsf{G},w_\mathsf{B})~=~(\mathtt{0.01}, \mathtt{0.04}, \mathtt{0.16})$ to approximate the nonlinear temperature differences between colors.

At calibration, we perform these intensity computations and obtain a baseline intensity value $I_\mathsf{cal}$ against by which all other intensities are scaled.
Let $\tilde{I} = I / I_\mathsf{cal}$ be the intensity expressed relative to this baseline.
This expresses the intensities with a physical meaning, for instance $\tilde{I} = 1$ is an intensity equal to that of the torch flame, while $\tilde{I} = 5$ is an intensity five times greater.
This simplifies the interpretation of intensity values and the tuning of desired intensity values.

To design a pool combustion state descriptor, the pool intensity must be normalized for combination with the pool convexity.
For this, a saturation intensity $\tilde{I}_\mathsf{sat}$ is set to be greater than the maximum values of $\tilde{I}$ observed during trial cutting runs.
In our experiments $\tilde{I}_\mathsf{sat} = 10$, \textit{i.e.},  intensity saturates at $10$ times the baseline.
We thus normalize the pool intensity as $i = \min (\tilde{I}_\mathsf{sat}, \tilde{I}) / \tilde{I}_\mathsf{sat}$ with $i \in (0, 1]$.

\subsection{Heat Pool Combustion State}\label{subsec:heat_pool_combustion_state}
Finally, the pool combustion state descriptor is defined as $s~=~\lambda c + (1 - \lambda) i$ where the relative emphasis on pool convexity or pool intensity is tuned with $\lambda\in [0, 1]$.
This yields a desirable normalized state descriptor $s \in (0, 1]$.
In our experiments, we set $\lambda = \frac{1}{2}$.
This descriptor, by design, captures the positive relationship between convexity and combustion state (whereby a higher convexity indicates more concentrated combustion), and the positive relationship between intensity and combustion state (whereby a higher intensity indicates a stronger combustion).

%% file: sections/6_Control.tex
The combustion control task assumes that calibration and conditioning are completed, meaning that the torch flame’s centroid and baseline intensity are determined (so that combustion state $s$ is computable) and that the metal surface is sufficiently preheated for combustion to take place.

\subsection{Modeling Assertions}

The controller is designed under the following assertions:

\subsubsection{The pose transform between the torch flame and camera is fixed}
The camera is rigidly attached to the torch, observing its tip from a fixed viewpoint throughout cutting and keeping the torch flame stationary in the image frame.
After calibration, the torch flame's centroid in the image frame does not change.

\subsubsection{The pool combustion state and torch speed have a negative relationship}
The pool combustion state is highly-correlated with the pool's temperature.
A faster torch speed reduces the amount of heat transferred to the local metal surface, thereby yielding a lower pool temperature and therefore a lower pool combustion state.
This is also confirmed empirically using a cutting torch and metal plates.
The pool combustion state drops at higher torch speeds, and increases at lower ones.
Note that the combustion state is taken to be an instantaneous measure of the most immediate pool, and does not concern residual heated regions elsewhere on the surface.

\subsubsection{There exists a desired speed at which the desired combustion state is maintained}
This results from the negative relationship between combustion state and torch speed.
Moving the torch too fast produces a deficient pool combustion, yielding poor or no cuts. Conversely, moving too slow produces an excessive combustion, yielding inefficient and badly textured cuts.
There is a range of intermediate speed values yielding adequate combustion among which exists a desired pair for speed and combustion state.

\subsection{System Variables}
We formally define the variables for heat source’s tangential velocity and the heat pool combustion state  as follows:

\begin{itemize}
    \item $v(t) \in \mathbb{R}_{\geq 0}$ is the tangential velocity of the heat source (torch flame centroid) on the reference cutting path.
	\item $s(t) = \phi(v(t)) \in \mathbb{R}_{> 0}$ is the pool combustion state in the image frame, where $\phi$ models the map from $v$ to $s$.
	\item $s^*$ is the desired state and $v^*$ is the desired velocity. 
\end{itemize}

The function $\phi: v \mapsto s$ maps from the torch velocity to the combustion state and thus admits the following conditions:

\begin{enumerate}
	\item $\phi$  is strictly positive: ${\phi(v) > 0, \forall v \in \mathbb{R}_{\geq 0}}$. This is since $s>0$ due to $c,i>0$ as defined in Section~\ref{subsec:heat_pool_combustion_state}.
	\item $\phi$ is monotonically decreasing with respect to $v$:
	$\tfrac{\mathrm{d}}{\mathrm{d}v}\phi(v)~<~0,\; \forall v(t)$ such that $\displaystyle \lim_{v \to \infty} \phi(v) = 0$. This is in accordance with the second modeling assertion.
	\item $\phi$ admits the desired-velocity constraint $\phi(v^*) = s^*$ in accordance with the third modeling assertion.
\end{enumerate}
	
\subsection{Control Law and Stability Proof}
The control input is the tangential acceleration $\dot{v}(t)$ updating the velocity $v$ of which the state $s$ is a function.
We thus control combustion state via acceleration.
The implication of $\phi(v^*) = s^*$ is that by tracking $s^*$, the torch is moved at the desired velocity $v^*$.
Let $e_s (t) = s^* - s(v(t))$ be the combustion state error with respect to the desired state $s^*$.
Its dynamics are: $\dot{e}_s (t)=-\tfrac{\mathrm{d}} {\mathrm{d}v}s(v)\dot{v}(t)$.
Apply the control input ${\dot{v}(t) = -k e_s(t)}$ with $k>0$ being a strictly positive gain.
This results in the desired behavior of accelerating when $s$ is excessive and decelerating when $s$ is deficient.
Substitute the control input in the dynamics equation to obtain:
\begin{equation}
    \dot{e}_s(t) = ke_s(t) \tfrac{\mathrm{d}} {\mathrm{d}v} s(v)
\end{equation}

We prove stability using the Lyapunov candidate function $V(e_s )=\frac{1}{2} e_s^2$, computing its derivative:
\begin{equation}
    \dot{V}\left(e_s(t)\right) = e_s \dot{e}_s = k e_s^2(t) \tfrac{\mathrm{d}} {\mathrm{d}v}s(v) < 0, \quad \forall e_s(t) \neq 0
\end{equation}

The above result is true for all non-zero error states since $k>0$, and $e_s^2 (t)>0$, and $\tfrac{\mathrm{d}} {\mathrm{d}v}s(v) = \tfrac{\mathrm{d}} {\mathrm{d}v}\phi(v) <0$. The controlled system is thus asymptotically stable.

We note that the desired combustion state $s^*$ is determined experimentally.
Specifically, an adequate range of combustion states is found for a particular setup and would not change across cutting sessions.
After a certain number of trials, a precise value for the desired state $s^*$ is determined. 

%% file: sections/7_Evaluation.tex
\begin{figure}[!t]
  \begin{center}
  \includegraphics[width=0.48\textwidth]{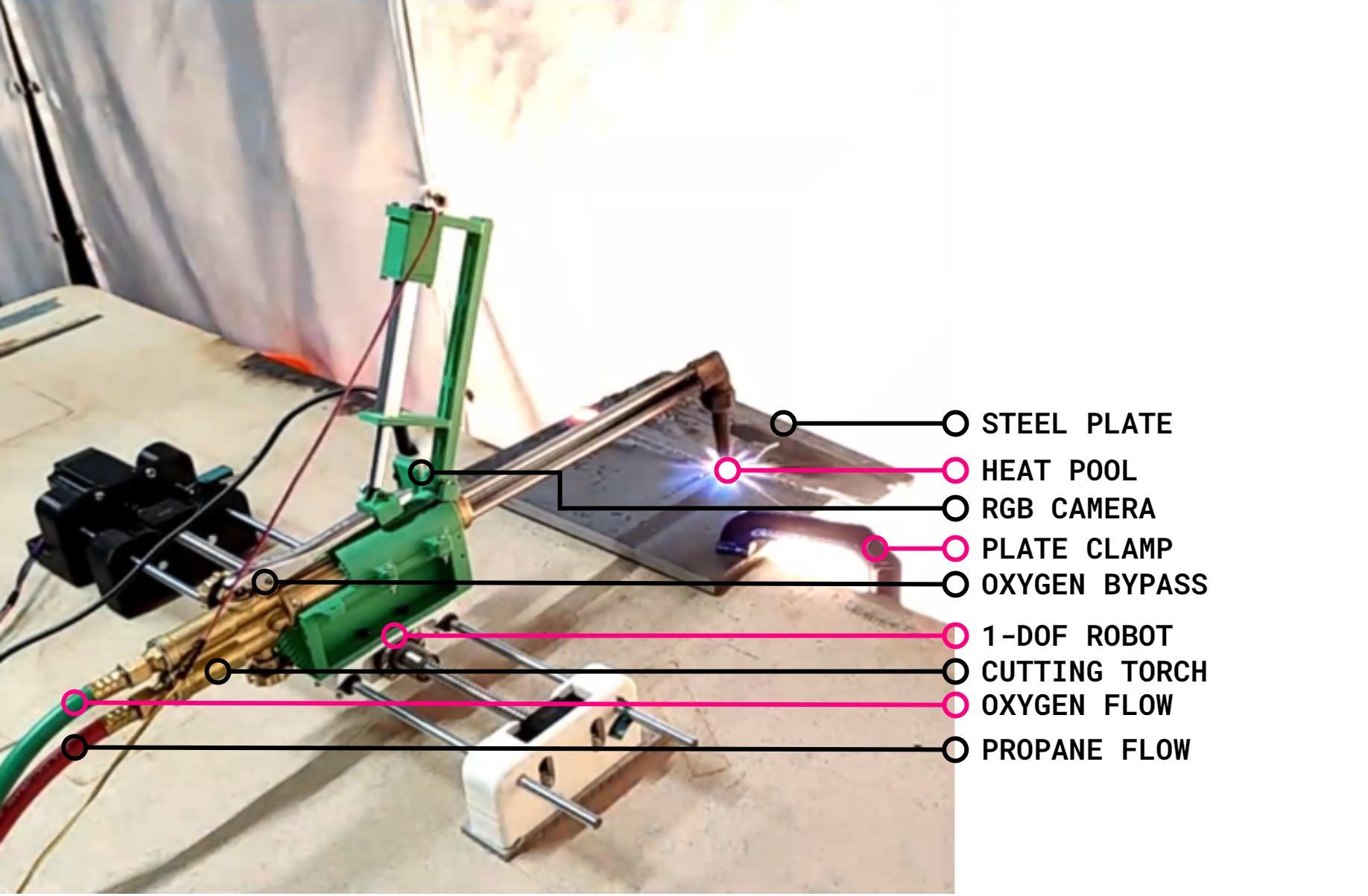}
  \end{center}
  \caption{Steel plates of varying thicknesses are clamped and secured before cutting.
  The 1-DOF robot equipped with its camera and cutting torch is installed, connected to oxy-propane sources and to its computer communications.
  The system can then receive feedback from the camera to perform the aforementioned vision-based control and cut the plate.
  }
  \label{fig:setup}
\end{figure}

We evaluate our vision-based cutting control algorithm in physical cutting experiments performed using a 1-DOF robot, a cutting torch, oxy-propane gas, and steel plates.
The physical setup is shown and annotated in Fig.~\ref{fig:setup}.
In our robotic setup, the camera is mounted at a fixed pose towards the torch tip.
The camera lens is covered with a tinted visor used to dim the scene and focus on the flame, as is typically done by skilled cutters.
This also prevents image saturation due to the extreme brightness of the flame and pool.
The actuators are a stepper motor setting the torch's linear motion, and a linear actuator engaging the oxygen bypass. 
Using ROS~\cite{Quigley2009}, the system communications are distributed across a microcontroller for the low-level actuator control and a computer for image processing and high-level control actions. 
The computer is connected to the camera and the microcontroller, while the torch is connected to oxygen and propane tanks. 

\subsection{Experiment Workflow}
The experiment workflow follows the oxy-fuel cutting sequence in Fig.~\ref{fig:cutting_timeline}.
After setup, the experiment begins by igniting the cutting torch and setting the oxy-propane pressures; these are done manually.
When a proper torch flame is achieved, calibration is started and the robot identifies the torch flame's centroid and its baseline intensity.
Next, the robot is commanded to move the torch flame to an initial location on the metal plate to preheat the surface.
Once the surface is sufficiently preheated (determined by the experimenter), the robot is entered into autonomous combustion control.
The robot engages the oxygen bypass and autonomously sets the torch speed based on its vision-based control.
In these experiments, the reference cutting paths for the 1-DOF robot are straight lines along the plate's surface.
After cutting, the bypass is disengaged and the torch is turned off.

\subsection{Experiment Results}
\begin{figure}[!t]
  \begin{center}
  \includegraphics[width=0.48\textwidth]{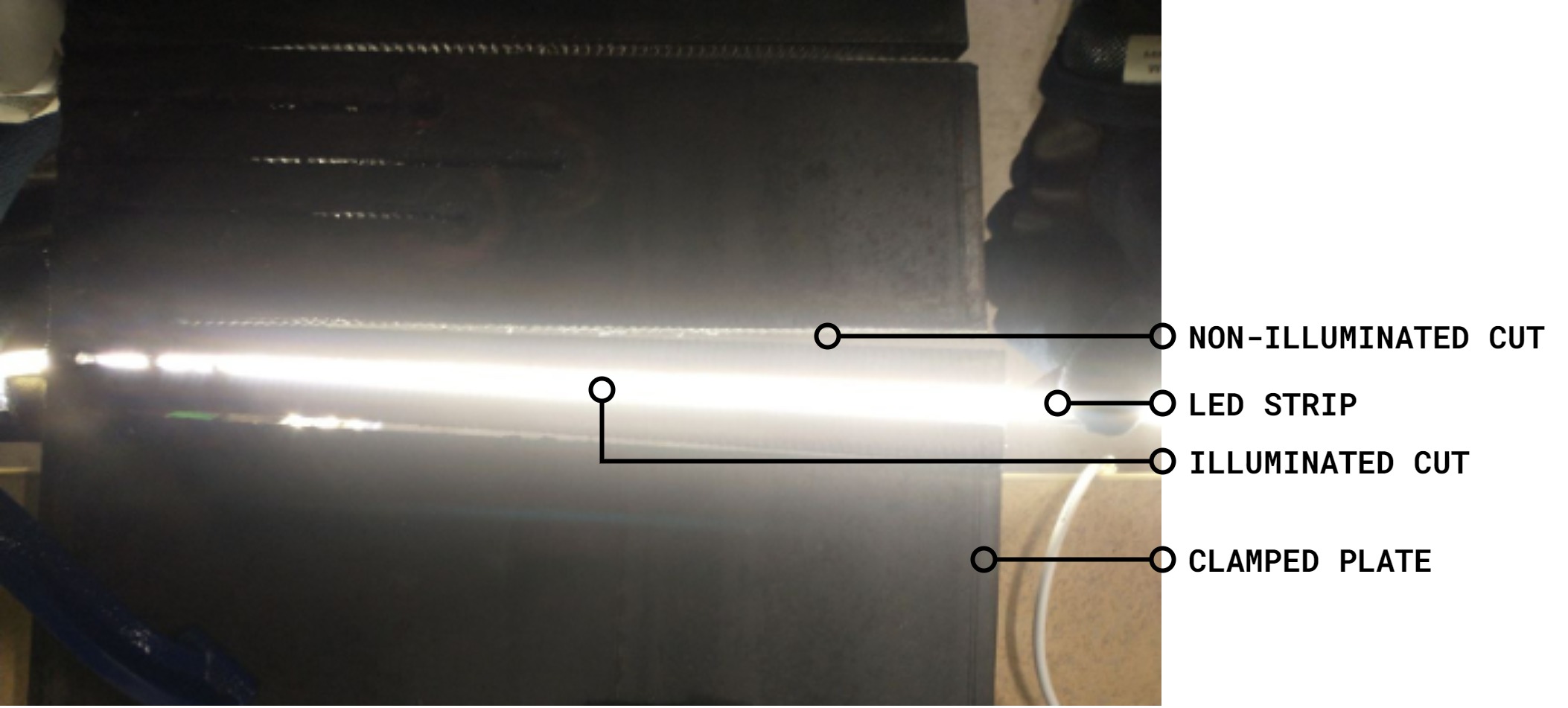}
  \end{center}
  \caption{After a cutting experiment, a LED strip is placed behind the plate and along the cutting path to inspect the cut and its overall quality.}
  \label{fig:led}
\end{figure}

The experiments are performed on steel plates of thicknesses $(\mathtt{0.250}, \mathtt{0.375}, \mathtt{0.500})\,\mathrm{in.}$.
We conducted experiments with three cutting modes to compare the outcomes of our automated vision-based algorithm against that of constant torch speeds:
(1) a constant torch speed of $\mathtt{0.2}\,\mathrm{cm} / \mathrm{s}$ (referred to as ‘slow cuts’);
(2) a constant torch speed of $\mathtt{3.2}\,\mathrm{cm} / \mathrm{s}$ (referred to as ‘fast cuts’); and, 
(3) controlled speed determined by the algorithm.
After the cut is performed on each plate by each mode, the plate is cooled down and each cut is illuminated for clear inspection by placing a LED strip behind the plate along the cutting path (see Fig.~\ref{fig:led}).

\begin{figure}[!t]
  \begin{center}
  \includegraphics[width=0.48\textwidth]{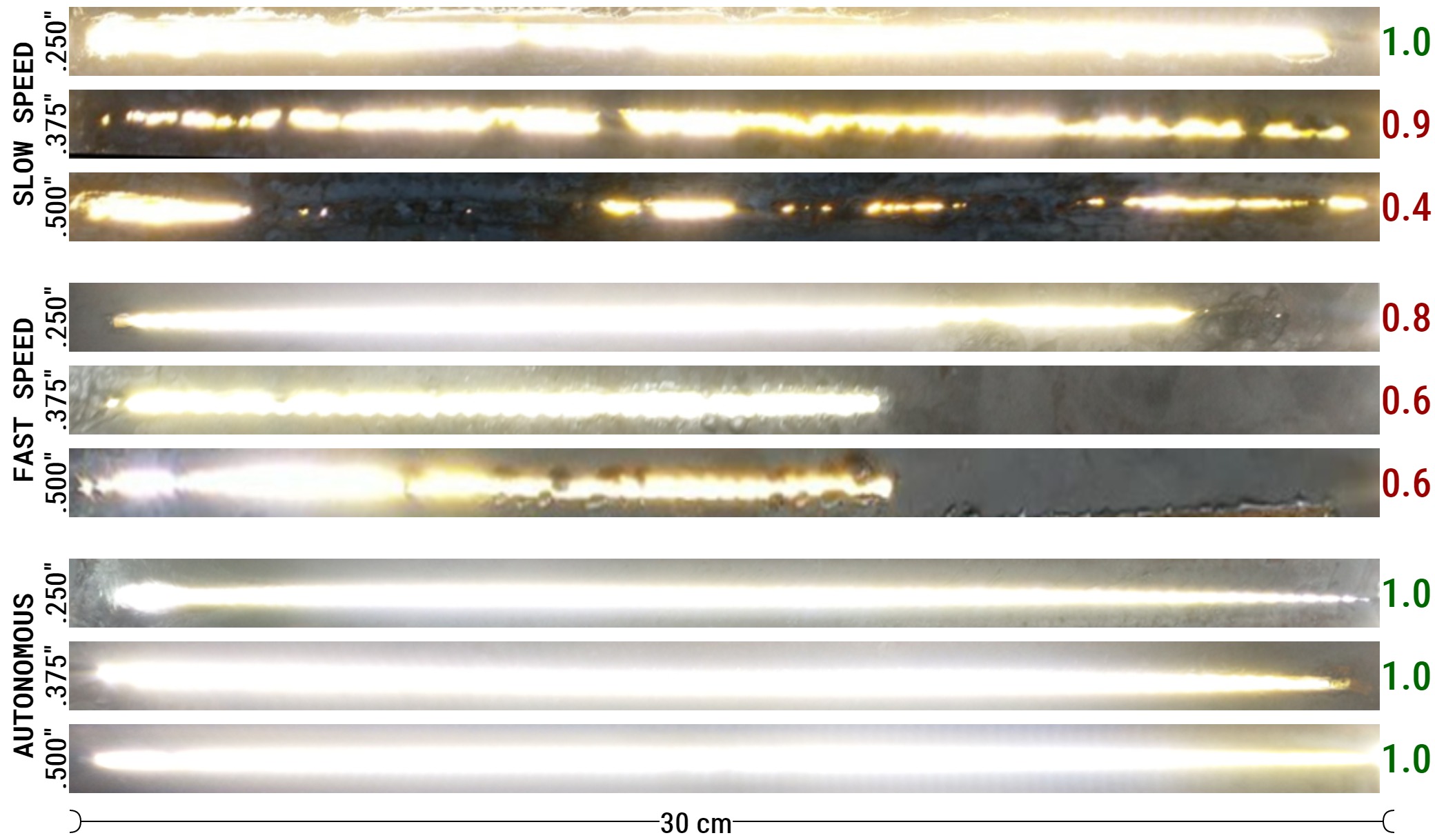}
  \end{center}
  \caption{The illuminated cuts of each mode (right) for each plate thickness (left), and the reference path length (bottom).}
  \label{fig:cuts}
 \vspace{-12pt}
\end{figure}

\begin{figure*}[!htb]
     \centering
     \begin{subfigure}[b]{0.329\textwidth}
         \centering
         \includegraphics[width=\textwidth]{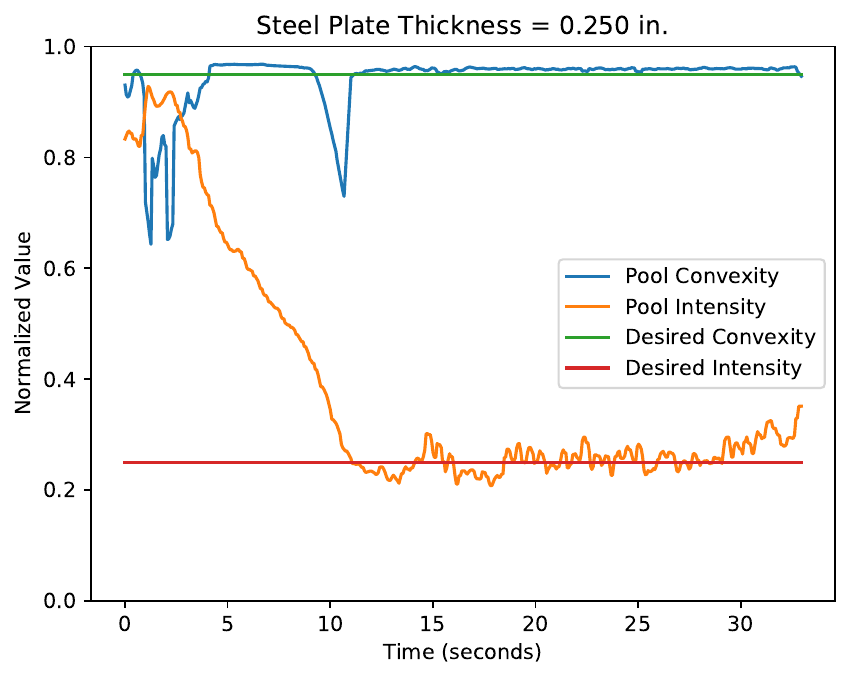}
     \end{subfigure}
     \begin{subfigure}[b]{0.329\textwidth}
         \centering
         \includegraphics[width=\textwidth]{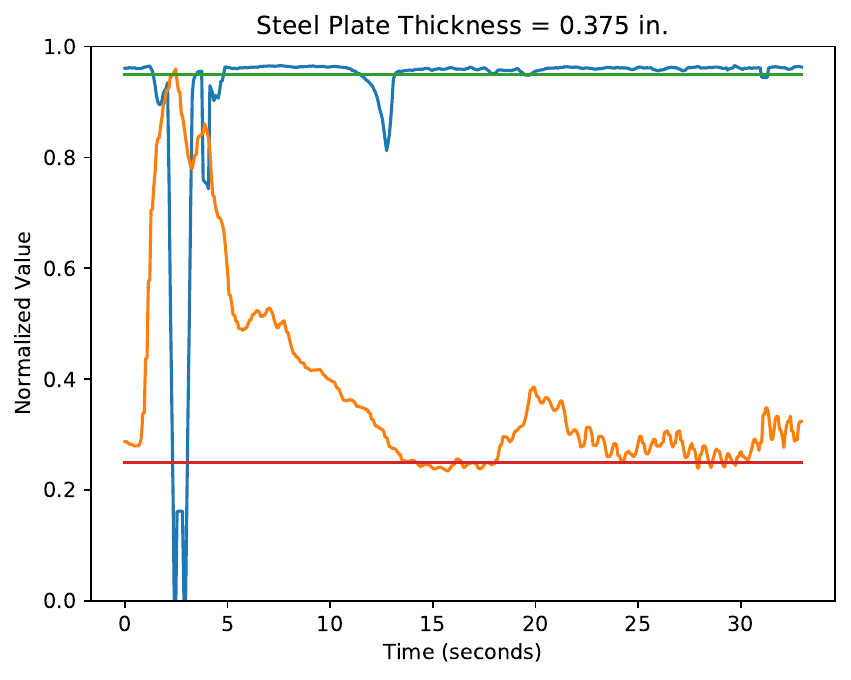}
     \end{subfigure}
     \begin{subfigure}[b]{0.329\textwidth}
         \centering
         \includegraphics[width=\textwidth]{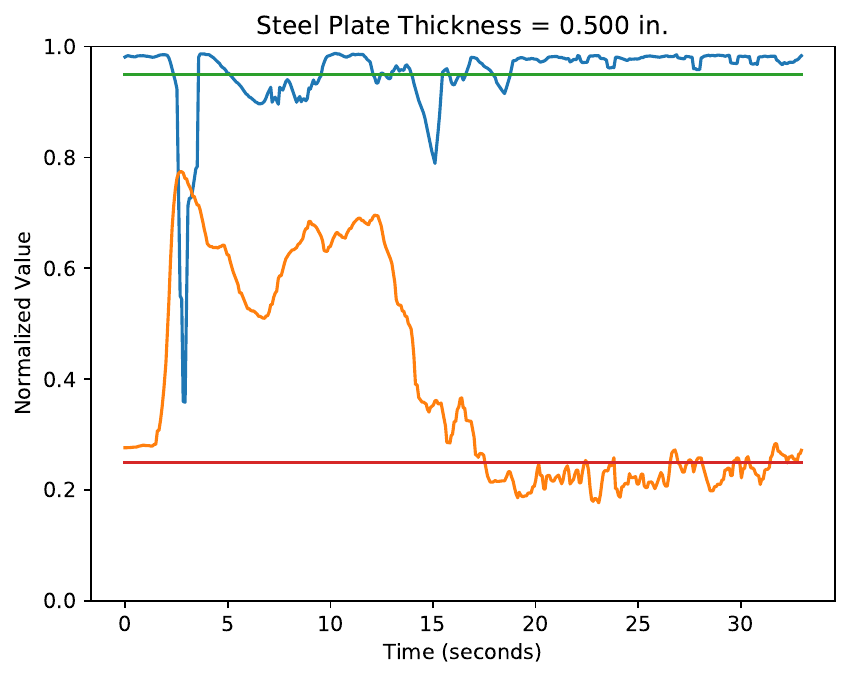}
     \end{subfigure}
     
     \centering
     \begin{subfigure}[b]{0.329\textwidth}
         \centering
         \includegraphics[width=\textwidth]{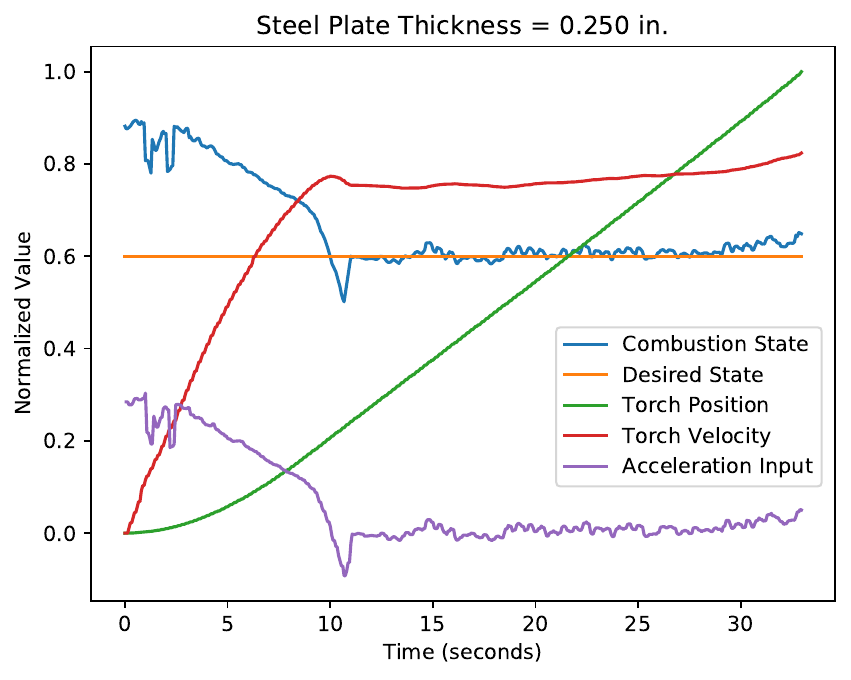}
     \end{subfigure}
     \begin{subfigure}[b]{0.329\textwidth}
         \centering
         \includegraphics[width=\textwidth]{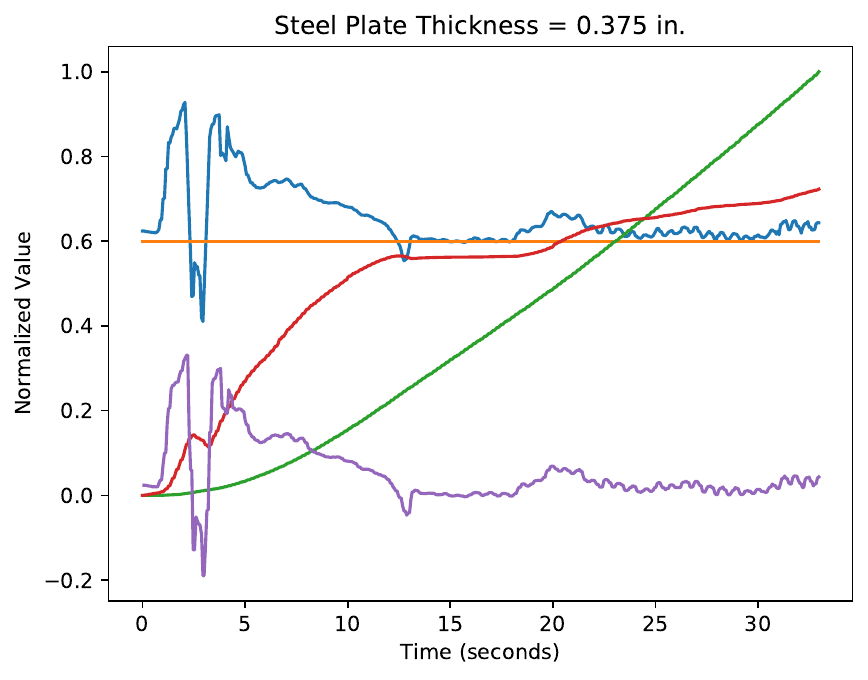}
     \end{subfigure}
     \begin{subfigure}[b]{0.329\textwidth}
         \centering
         \includegraphics[width=\textwidth]{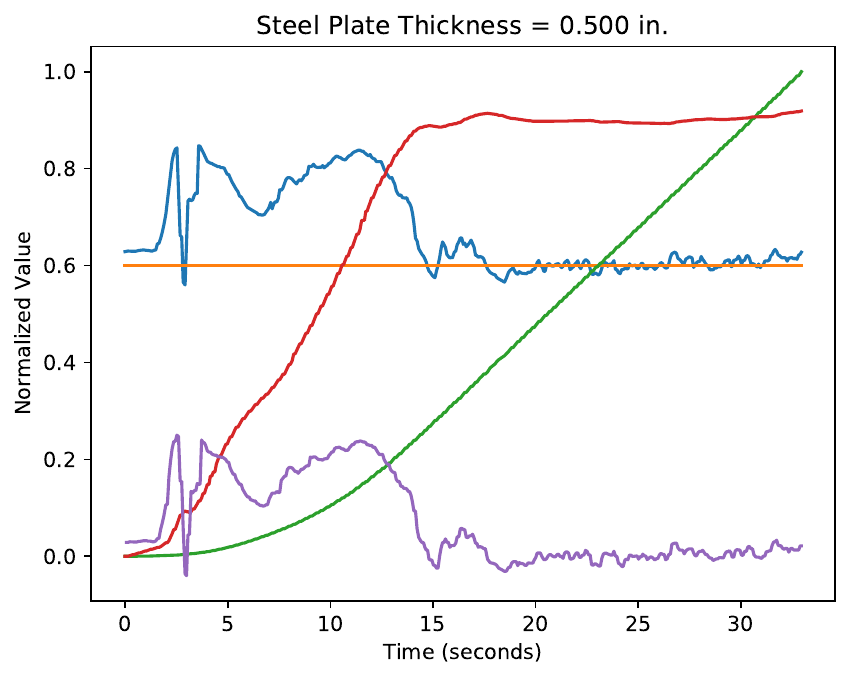}
     \end{subfigure}
  \caption{Time-series plots of the autonomous experiments (combustion cutting phase) for each plate.
  The top row shows both pool features, whereas the bottom row combines them into combustion state.
  The legends for each row are on the left.}
  \label{fig:plots}
 \vspace{-12pt}
\end{figure*}

In addition, the controller's performance is examined in time-series plots (Fig.~\ref{fig:plots}) wherein the following quantities are logged:
(1) torch position, (2) torch velocity, (3) acceleration input, (4) pool convexity and intensity (and when combined, combustion state), and (5) their desired values.
These are all plotted with normalized values.
Pool convexity, intensity, and combustion state are normalized by default.
Position is normalized by its maximum value reached at the cutting path's end.
Velocity and acceleration are normalized by their maximum values $v_\mathsf{max}$ and $\dot{v}_\mathsf{max}$ imposed as safety constraints on the controller; though these values are never attained.

All autonomous cuts are performed with these parameters: gain $k = \mathtt{200}$, pool convexity reference $c^* = \mathtt{0.95}$, pool intensity reference $i^* = \mathtt{0.25}$, maximum velocity $v_\mathsf{max}~=~\mathtt{2.0}\,\mathrm{cm} / \mathrm{s}$, maximum acceleration $\dot{v}_\mathsf{max}~=~\mathtt{0.8}\,\mathrm{cm} / \mathrm{s}^2$.
While the problem is noisy, the controller converges in every case and successfully tracks the desired combustion state.

Referring to Fig.~\ref{fig:cuts}, one may intuitively expect the slow cuts to fully penetrate each plate along the cutting path.
In actuality, moving the torch too slowly accumulates heat excessively in regions surrounding the cutting path thereby re-sealing portions of the cut with creeping deposits of heated steel (cut success ratios $\mathtt{0.4}$ and $\mathtt{0.9}$).
Note that the $\mathtt{0.250}\,\mathrm{in.}$ plate is thin enough that the excess heat combusts these surrounding regions yielding the wide cut (ratio $\mathtt{1.0}$).
Nevertheless, slow cuts are wasteful in time and resources in addition to yielding substandard cut textures.
The fast cuts instead tend to extinguish the heat pool before the cut's completion (ratios $\mathtt{0.6}$ and $\mathtt{0.8}$) due to diminishing heat accumulation, fault of moving the torch too quickly.
The effects of moving too slow or too fast are aggravated on thicker plates due to a larger material volume surrounding the cutting path.
In our vision-based formulation, these faults translate to combustion states that are excessive (slow cuts) or deficient (fast cuts).

By contrast, the autonomous cuts are clear and successful in every case.
Without knowing the plate thicknesses, the controller adjusts the torch velocity towards a stable region to converge and maintain the combustion state near its desired value (see Fig.~\ref{fig:plots}).
In effect, Fig.~\ref{fig:cuts} shows the resulting successful cuts against each of the plate thicknesses.

%% file: sections/8_Conclusion.tex
We formalize the oxy-fuel cutting problem and develop a vision-based controller for combustion cutting.
The vision feedback describes the heat pool's combustion state using its convexity and intensity, which the controller uses to update the torch velocity along the cutting path.
Our experiments demonstrate our controller's success on a 1-DOF robot in cutting steel plates of varying thicknesses using purely visual feedback.
While we solve the calibration and control tasks, further work is needed to automate surface conditioning for autonomous transitioning from preheating to cutting control.